\documentclass[letterpaper, 10 pt, conference]{ieeeconf}  

\IEEEoverridecommandlockouts                             
\title{\LARGE \bf
1000 Rallies: An Event-Camera Dataset and \\ Real-Time Learned Ball-State Estimation for Robotic Table Tennis}

\usepackage{graphicx}%
\usepackage{multirow}%
\usepackage{amsmath,amssymb,amsfonts}%
\usepackage{mathrsfs}%
\usepackage{textcomp}%
\usepackage{manyfoot}%
\usepackage{booktabs}%
\usepackage{algorithm}%
\usepackage{algorithmicx}%
\usepackage{algpseudocode}%
\usepackage{listings}%
\usepackage{comment}
\usepackage{url}
\usepackage{tabularx}
\usepackage{threeparttable}
\usepackage{acronym}
\usepackage{amssymb}
\usepackage[table]{xcolor}
\usepackage{subcaption}

\usepackage[acronym]{glossaries}
\newacronym{evs}{EVS}{Event-based Vision Sensor}
\newacronym{aps}{APS}{Active Pixel Sensor}
\newacronym{cnn}{CNN}{Convolutional Neural Network}
\newacronym{fov}{FoV}{Field of View}
\newacronym{sae}{SAE}{Surface of Active Events}
\newacronym{yolo}{YOLO}{You Only Look Once}
\newacronym{roi}{RoI}{Region of Interest}
\newacronym{ekf}{EKF}{Extended Kalman Filter}
\newacronym{cma-es}{CMA-ES}{Covariance Matrix Adaptation Evolution Strategy}
\newacronym{mae}{MAE}{Mean Absolute Error}
\newacronym{iou}{IoU}{Intersection over Union}
\newacronym{snn}{SNN}{Spiking Neural Network}

\definecolor{apspink}{HTML}{FF69B4}  
\definecolor{evsgreen}{HTML}{90EE90}  

\begin{document}
\author{Raphaela Kreiser$^{*}$, Asude Aydin$^{*}$, Yin Bi, Claudio Fanconi$^{\dag}$, Peter D\"urr, Naoya Takahashi \\
Sony AI, Zurich%
\thanks{$^{*}$These authors contributed equally.}%
\thanks{$^{\dag}$This work was completed while Claudio Fanconi was an intern at Sony AI, Zurich.}%
}

\maketitle

\begin{abstract}
Robotic table tennis has emerged as a compelling benchmark for real-time robotic perception due to its fast ball dynamics and stringent timing requirements.
Accurate, high-frequency, and low-latency ball state estimation is critical for reliable trajectory prediction and timely control.
Traditional frame-based cameras face an inherent trade-off: low frame rates leave temporal blind spots that miss fast-moving objects and high frame rates raise data and computational cost.
Event cameras instead offer microsecond temporal resolution and, under sufficient illumination, remain largely free of motion blur even at high ball speeds.
However, the community lacks large-scale datasets to develop and benchmark event-based perception in realistic sports scenarios.
We address this gap by introducing the first large-scale event-camera dataset for table tennis, comprising over 1000 rallies from a diverse group of players ranging from amateurs to elite-level athletes.
Each recording captures the event stream alongside 14 synchronized high-speed frame-based cameras at 200~FPS, which we use to produce 1~kHz pseudo ground-truth labels for ball position, velocity, and spin.
Building on this dataset, we train a convolutional neural network robust to background player motion that jointly estimates the ball's position and velocity in the image-plane from events.
Treating the predicted velocity as an additional measurement in the Kalman filter reduces bounce-point prediction error by 36\% relative to a position-only baseline.
Finally, we close the perception–action loop by integrating the event-based system with a St\"aubli robotic arm, enabling the first real-time human–robot table tennis rallies driven by event-based perception.
\footnote{The dataset and code will be made publicly available.\label{fn_data_avaiability}}
\end{abstract}
    
\section{INTRODUCTION}

Robotic table tennis has long been regarded as a challenging benchmark for real-time perception and control, requiring accurate ball state estimation, reliable trajectory prediction, and high-speed motion planning under stringent latency constraints~\cite{hashimoto_learning_TT,d2024achieving,tebbe2019table}. 
In professional play, ball velocities can reach up to 35 m/s~\cite{ball_stats}, allowing the ball to traverse the table in less than 200 ms.
Accurate and timely trajectory prediction for competitive returns therefore depends on low-latency sensing and precise, high-frequency ball state estimation, including its position, velocity, and spin.
Recent advances have enabled robots to rally with amateur players under modified equipment and simplified rules~\cite{dambrosio2025achieving, su2025hitter, ma2025learning, wang2025integrating}. 
Only recently have Dürr et al.~\cite{durr2026outplaying} demonstrated table tennis rallies with elite athletes, underscoring the growing appetite for high-speed visual sensing in dynamic robotics.
The perception system used in that work achieves its speed through color-based segmentation of the orange ball, tightly coupled to a specific visual signature. 
Event cameras offer a complementary route: by encoding motion rather than appearance, they enable high-speed perception without relying on any particular color.


\begin{figure}[tbp]
    \centering
    \includegraphics[width=0.95\linewidth, trim=0 50pt 0 0, clip]{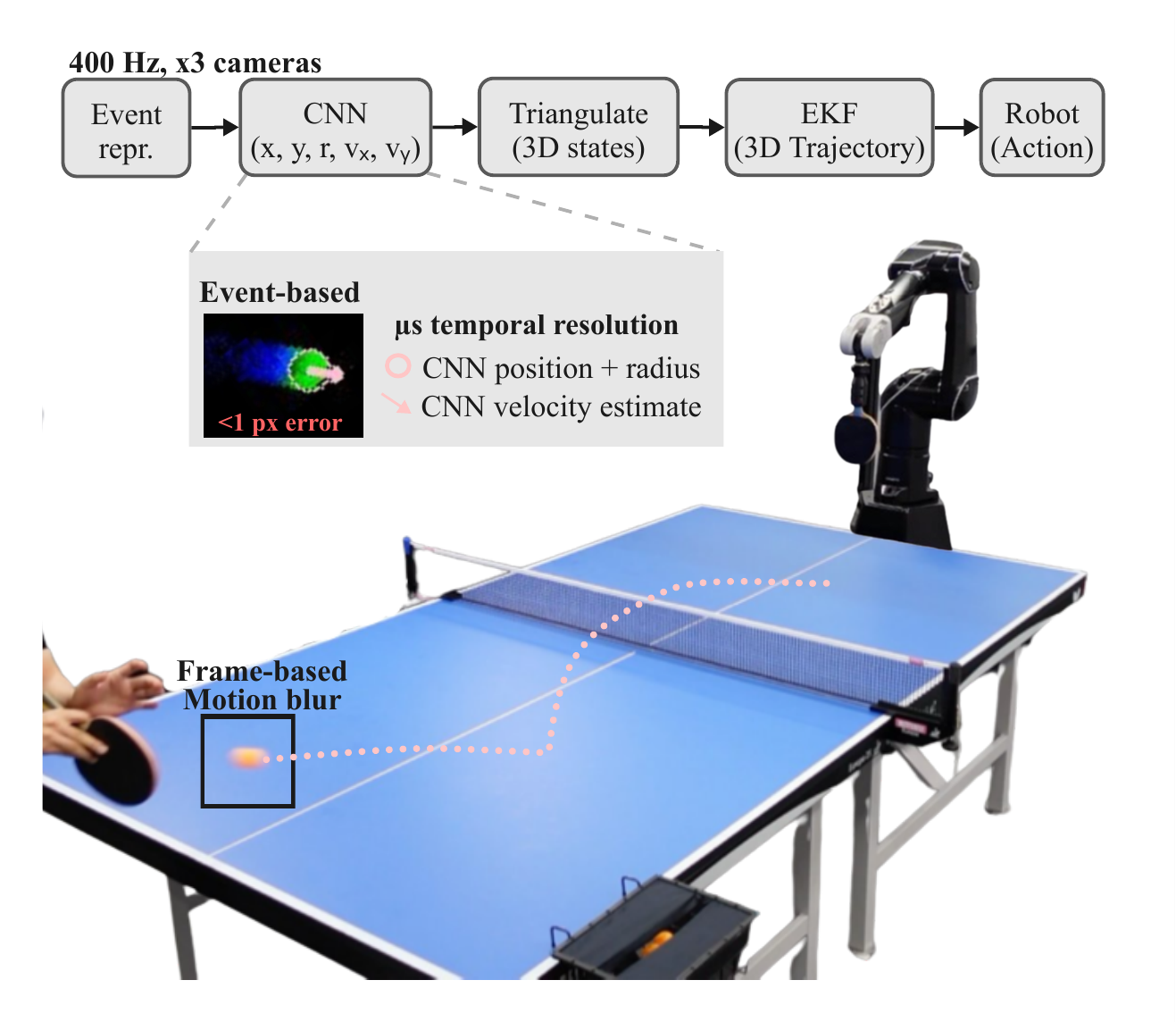}
    \caption{
    Three event cameras capture the playing area at high temporal resolution. Our CNN jointly predicts ball position and velocity from the event stream at 400 Hz, driving an EKF trajectory predictor that closes the perception–action loop and enables human–robot rallies.}
    \label{figures:robot_events}
\vspace{-15 pt}
\end{figure}

The \gls{evs}, a neuromorphic sensing technology, asynchronously reports pixel-level brightness changes at microsecond resolution rather than capturing full images at fixed intervals~\cite{gallego2020event}.
Previous works have explored \gls{evs}-based ball tracking for table tennis to leverage such properties for accurate and low-latency state estimation, demonstrating that higher-frequency state updates directly reduce the trajectory prediction error~\cite{ziegler2025event, alberico2025egocentric}. 
Alberico et al.~\cite{alberico2025egocentric} focus on an egocentric view, while Ziegler et al.~\cite{ziegler2025event} use fixed cameras; both approaches rely on conventional methods for ball detection, such as clustering or circle detection within a small \gls{roi}.
However, such methods often fail to generalize to practical scenarios due to their reliance on accurately defined \gls{roi}s, susceptibility to spurious background activity, and dependence on hyperparameters tailored to factors such as event rate, sensor bias settings, and environmental lighting conditions.

In this work, we first introduce a comprehensive dataset of five hours of \gls{evs} table tennis rally recordings ($>$1000 rallies, amateur to elite athletes with over ten years of competitive experience) collected from four different camera views in two different lab environments, capturing realistic ball-game occlusions and dynamic player motions.
We present a pipeline for generating 1~kHz ground-truth labels for the event cameras using \gls{aps} camera measurements upsampled from 200~Hz, providing the ball's position, size, and velocity in the image plane, as well as ball spin estimates obtained from the \gls{aps} cameras using a dotted-ball pattern~\cite{gossard2023spindoe}.
We envision that these recordings and the labeling pipeline address the community's need for developing and benchmarking event camera-based perception in realistic sports scenarios involving fast-moving balls.

Building on this dataset, we propose a \gls{cnn}-based approach for accurate and robust ball state estimation, 
enabling, to the best of our knowledge, human–robot table tennis rallies with event cameras for the first time.
In addition to predicting the ball’s size and location on the image plane, as in standard circular object detectors, we also estimate its image plane velocity directly from the event representation in a single-shot manner.
This enables faster and more accurate trajectory prediction, eliminating the need for longer sequences of ball detections.
Concretely, supplying the predicted image-plane velocity as an additional \gls{ekf} measurement reduces 100~ms bounce-prediction error by 36\% (12.1~cm $\rightarrow$ 7.7~cm) over a position-only baseline.
We develop a highly optimized C++ implementation that delivers real-time performance on commodity hardware.
This implementation drives detections with three event cameras at $1280\times720$ resolution with update rates of 400~Hz.

Leveraging the proposed \gls{cnn} approach and the richness of the training dataset, our method achieves state-of-the-art accuracy in event-based table tennis ball detection.
Finally, we close the perception-action loop by integrating our event-based ball detector pipeline into a robotic framework through~\gls{ekf} trajectory predictions, as illustrated in Fig.~\ref{figures:robot_events}.
Our contributions are summarized as follows:

\begin{itemize}
    \item We introduce the first large-scale event-camera dataset for table tennis (over 1000 rallies from amateurs to elite-level athletes), together with a labeling pipeline that produces upsampled pseudo ground-truth labels at 1~kHz, including ball position, velocity, and spin$^{\ref{fn_data_avaiability}}$.
    \item We propose a data-driven \gls{cnn} for event-based ball-state (position and velocity) estimation in the image plane, achieving state-of-the-art accuracy on \gls{evs}-based table tennis which can track professional-level high-speed shots and generalize to unseen environments.
    \item To the best of our knowledge, we introduce the first event-based perception system fully integrated into a robotic setup, enabling sustained human–robot rallies through an optimized C++ implementation that runs three cameras at a 400~Hz detection rate.
\end{itemize}

\section{Related work}

\subsection{Event-Based Ball Detection}

Tracking a fast-moving, small object such as a table tennis ball
poses significant challenges related to capture rate, exposure time, and its limited size within the field of view.
Frame-based vision systems have achieved real-time ball trajectory prediction \cite{voeikov2020ttnet, gomez2020real}, although detection rates are often limited either by camera throughput \cite{tebbe2019table} or by the computational cost of processing high frame-rate data. 
In contrast, event cameras with their microsecond temporal resolution and asynchronous operating principle \cite{gallego2020event}, provide a compelling alternative by reducing redundant data and suffering less from motion blur compared to conventional APS cameras.
Learning-based, event-driven general object trackers such as SpikeMOT~\cite{wang2024spikemot}, EV-Tracker~\cite{zhang2022evtracker}, and Perez et al.~\cite{perez2022event} have demonstrated robust, motion-blur-resistant tracking in dynamic environments. 
However, these general-purpose trackers are not tailored to small, fast-moving objects such as balls
and operate at update rates lower than 100 Hz.

More targeted event-based ball detectors tailored to table tennis have pushed update rates considerably higher by combining \gls{roi} cropping with classical detection methods.
Ziegler et al.~\cite{ziegler2025event} achieve ball detection rates of $4.14 \times 10^3$~Hz using a fixed \gls{roi} combined with Hough-based circle detection, and demonstrate that higher update rates improve the accuracy of downstream filter-based trajectory prediction.
Similarly, Alberico et al.~\cite{alberico2025egocentric} achieve update rates of up to 200~Hz using a dynamically adjusted \gls{roi} guided by an eye-tracking module in an egocentric AR setup, where ball detection is performed via DBSCAN clustering~\cite{ester1996density} followed by circle fitting.
While \gls{roi}-based classical detectors achieve high update rates, they are susceptible to failure under occlusion \cite{ziegler2025event} or gaze disruption \cite{alberico2025egocentric}. 
Moreover, as purely hand-crafted methods, they lack the ability to learn from data or adapt to new environments without manual re-tuning. 

Several recent works have applied neural networks to event-based ball detection in sports settings.
Nakabayashi et al.~\cite{nakabayashi2023event} proposed the first deep-learning-based ball detector for event cameras, trained on synthetic events and demonstrated detection of balls invisible in conventional frames due to motion blur.
EV-Catcher~\cite{wang2022ev} combined a lightweight learned event representation with a neural network to predict the ball's 1D horizontal impact location and time-to-collision for robotic catching at speeds up to 13~m/s.
While these learning-based approaches advance event-based ball perception, they either target only 1D or 2D detections without velocity estimation or operate at limited update rates, posing difficulties for real-time applications with faster ball dynamics.

\subsection{Event-based Robotic Applications}
Several works have explored integrating event cameras into dynamic robotic tasks. Ball-catching systems such as Forrai et al. \cite{forrai2023event} and EV-Catcher~\cite{wang2022ev} represent important steps toward event-driven robotic interception, but rely on a single camera and a cropped \gls{roi} for more constrained ball-catching scenarios. 
The closest to our application scenario is that of Ziegler et al.~\cite{ziegler2025event}, which targets human–robot table tennis but has not yet been integrated into a complete robotic system.
Complementary efforts have explored neuromorphic and spike-based sensing for table tennis robots. 
Ziegler et al.~\cite{ziegler2025detection} benchmarked \glspl{snn} on neuromorphic edge devices for event-based ball detection, demonstrating real-time inference at a fraction of GPU power consumption, though at reduced accuracy compared to NN-based approaches. 
SpikePingpong~\cite{wang2025spikepingpong} employed a 20~kHz spike camera—distinct from event cameras in its operating principle—with learned trajectory correction while relying on \gls{aps} for ball detection. 
Both of these works integrate their respective perception pipelines into physical robotic systems, but are evaluated with ball launchers under controlled conditions. 
In our work, we demonstrate full pipeline integration, with perception driving a motion planner based on Ruckig \cite{berscheid2021jerk} that generates jerk-constrained trajectories to enable rally play against human opponents.

\subsection{Event Camera Datasets}
Many event camera datasets have been proposed for object detection and tracking, for example in automotive scenarios \cite{perot2020learning}
human actions and gestures \cite{hu2016dvs}, people detection \cite{boretti2023pedro}, and general object categories \cite{wang2024event}.
However, these datasets are domain-specific and not suitable for high-speed ball tracking in sports.
Recently, two concurrent works have introduced event-based table tennis data from an overhead camera setup; however, neither dataset is publicly available at the time of submission. 
Ziegler et al.~\cite{ziegler2025event} provide only four short ball trajectories in an otherwise static scene, with no background activity from players. 
In contrast, Wang et al. collect data across several controlled settings using two cameras, but report ball speeds only for simulated data (5--9~m/s), without providing velocity or player skill information for real-world recordings.
Both works use a stereo camera setup to collect data; however, they rely on manual annotation to obtain 2D ball positions in the image plane. 
This limits the annotations to image-based positions only, without 3D trajectory, velocity, or spin information, and constrains the dataset scale due to the required labeling effort.
In this work, we introduce a large-scale event camera dataset collected from competitive table tennis matches, featuring players ranging from amateur to elite level and ball speeds of up to 26~m/s. 
Our dataset comprises approximately 1,200 rallies captured by four synchronized event cameras distributed around the table for broader spatial coverage with 3D position, velocity, and spin annotations obtained from a calibrated multi-camera \gls{aps} system of up to 14 cameras.

\section{Dataset}

\begin{table*}[t]
\centering
\caption{Statistics of the recorded data. The shaded row indicates data collected from a different lab.}
\label{tab:dataset_stats}
\begin{tabular}{@{}l|ccccccc@{}}
\toprule
\textbf{Type} &
  \textbf{Duration (min)} &
  \textbf{\# Cameras} &
  \textbf{Mean Rate (Mev/s)} &
  \textbf{Max Rate (Mev/s)} &
  \textbf{\# Players} &
  \textbf{\# Rallies} &
  \textbf{Max Ball Speed (m/s)} \\ \midrule
\textbf{Elite} & 276.5 & 4 & 0.8 & 1.5 & 7 & 1167 & 25.9 \\
\textbf{Robot}      & 29.0  & 4 & 0.6 & 0.9 & 2 & 35  & 6.5    \\
\rowcolor[HTML]{EFEFEF} 
\textbf{Amateur}    & 5.3   & 3 & 1.4 & 2   & 2 & 27  & 6.5    \\ \bottomrule
\end{tabular}
\end{table*}

To enable the data-driven approach for ball state estimation, we collect and curate a dataset of EVS recordings from real-world table tennis matches.

\subsection{Data Collection}

We recorded 5\,h\,10\,min of table tennis data consisting of approximately 1200 rallies with 11 unique players ranging from amateur players to elite-level athletes (see Table~\ref{tab:dataset_stats}).\footnote{All table tennis players recorded in this dataset provided informed consent, and the data was collected in accordance with our organization's privacy and ethical regulations.}
During recording sessions, players are paired and instructed to play competitive matches, actively attempting to outscore their opponent and vary their tactics as they would in real tournament conditions.
This allowed us to cover a wide range of shots in terms of ball placement, speed, rally length, and play area.

\begin{figure*}[t]
  \centering
  \includegraphics[width=0.9\textwidth]{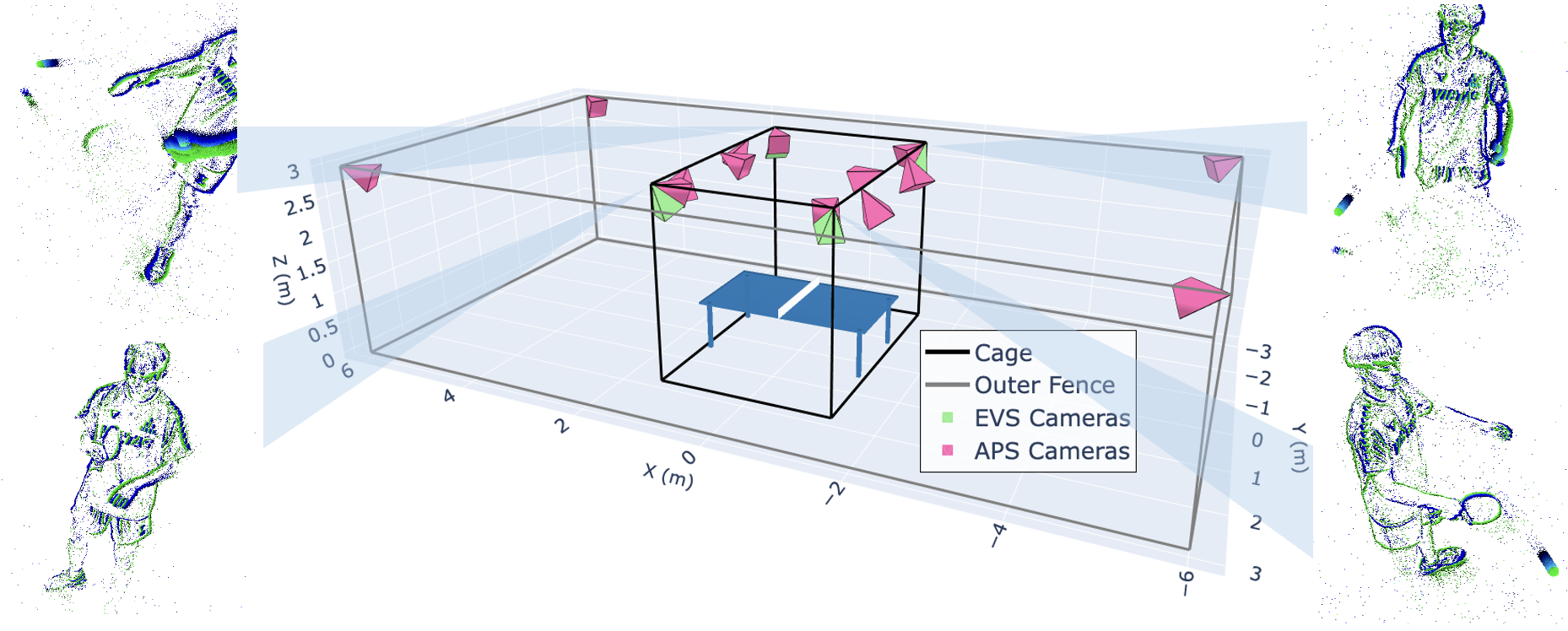}
  \caption{3D schematic of the multi-camera setup for table tennis data collection, featuring four event cameras \textcolor{evsgreen}{(green)} at the inner cage and distributed \gls{aps} cameras \textcolor{apspink}{(pink)} around inner and outer cages for ground-truth collection. Camera viewing directions are roughly indicated by pyramid orientations with example visualizations of the event stream per \gls{evs} view.}
  \label{fig:dataset_camera_setup}
\end{figure*}

For conducting the recordings, four event cameras (Prophesee EVK4 equipped with Sony IMX636 HD-resolution sensors) are mounted at the top corners of the inner cage, pointing downward toward the table.
The event cameras are calibrated to 14 frame-based \gls{aps} cameras with Sony IMX273 sensors operating at 200 Hz, which were placed around the inner and outer cages to cover the entire playing area, as seen in Figure~\ref{fig:dataset_camera_setup}. All recordings used the same lab and four-EVS / fourteen-APS configuration except for the amateur rallies (shaded row in Table~\ref{tab:dataset_stats}), which were collected in a different lab with three-EVS / six-APS cameras to evaluate generalization. For all labs, camera calibration is done for overlapping camera clusters using the Kalibr calibration toolbox \cite{furgale2013unified} and its event-based adaptation proposed by \cite{muglikar2021calibrate}.
The APS and event cameras, as well as the robot's actuators, are synchronized with a 200~Hz trigger signal.

\subsection{Data Annotation}

\textbf{APS Ball Detection}~The multi-camera APS system acts as an external ball position measurement system for ground truth generation in the EVS image plane. Each APS camera captures Bayer-pattern images that are converted to BGR, Gaussian-blurred, and transformed to HSV space for illumination-robust segmentation. In particular we use orange table tennis balls for color-based masking. 
Additionally, a temporal motion filter suppresses stationary false positives by differencing grayscale frames and intersecting the resulting motion mask with the color mask. Connected components are then extracted, and contours are evaluated based on circularity.
Valid detections are obtained by fitting minimum enclosing circles and retaining only those within a plausible range given the focal length and radius of the ball.
\\

\textbf{APS Spin Estimation}
\label{sec:spin_estimation}

To estimate the ball's spin, we employ a dotted-ball orientation estimation method adapted from SpinDOE~\cite{gossard2023spindoe}.
In this approach, a known dot pattern is applied to the ball's surface using a 3D-printed stencil. The dots are detected in the \gls{aps} camera images, and their spatial configuration is matched to the reference pattern using geometric hashing. The ball's orientation is then recovered via the Kabsch algorithm, which computes the optimal rotation between the detected and reference dot positions.
By estimating the ball's orientation from consecutive \gls{aps} frames, the spin is computed as the angular velocity between successive orientations.
The spin vector $\boldsymbol{\omega}(t)$ is decomposed into its rotation axis and magnitude using a quaternion-based regression~\cite{tebbe2020spin}, providing spin estimates at the \gls{aps} frame rate of 200~Hz, which are then used as measurements for the \gls{ekf}.
\\

\textbf{Label Generation}~Ball contact events - such as table hits, net hits, and racket contacts - are hand-labeled to segment free-flying trajectories. 
These labels are derived from triangulated 3D ball positions $\mathbf{p}_3(t) \in \mathbb{R}^3$, which have an average position error of 3 mm.
The subdivided free-flight trajectories, each representing a continuous observation period, are fitted to a third degree polynomial $k \in \{x, y, z\}$. $\mathbf{p}_{3,k}(t) \approx \sum_{j=0}^{d} c_{k,j} (t - t_1)^j$, where the coefficients $\{c_{k,j}\}_{j=0}^d$ minimize the least-squares error.
We empirically evaluated different polynomial orders and found the third degree to offer the best trade-off between fit accuracy and smoothness, with higher orders overfitting triangulation noise and lower orders failing to capture aerodynamic drag.
Since cross-axis coupling from drag and spin is negligible at this scale we fit the coefficients for each spatial dimension independently and validate that the per-axis fit residuals are below 1 mm in 3D, well within APS triangulation noise.
Using the fitted polynomials, we upsample the positions to 1 kHz, allowing for filling temporal gaps due to missing observations, as well as denser and smoother position and velocity estimates.
Velocities are then computed using backward finite differences on the upsampled trajectories, $\tilde{\mathbf{v}}_3(t_i) = (\tilde{\mathbf{p}}_3(t_i) - \tilde{\mathbf{p}}_3(t_{i-1}))/\Delta t$ for $i \geq 2$. In addition, ball spin is estimated from the \gls{aps} images using the dotted-ball method described above at 200~Hz and included as part of the dataset annotations.

Finally, processed 3D ball positions and velocities are back-projected into each event camera's image plane, which serve as ground-truth labels for the event cameras.


\section{Methods}

Robotic table tennis requires accurate, high-rate 3D ball-state estimation despite clutter, occlusions, and extreme speeds. 
Our pipeline pairs a \gls{cnn} detector, predicting ball position and image-plane velocity from \gls{evs} data, with an \gls{ekf} that fuses these through a physics-based motion model into 3D states and trajectory predictions for control.

\subsection{EVS Ball Detection}

We adapt the YOLO v4-tiny~\cite{bochkovskiy2020yolov4} model where the network takes in event representations and predicts ball location $(x, y)$, radius $(r)$, velocity $(v_x, v_y)$, and confidence.
While standard YOLO models parameterize objects using bounding boxes (width and height), we instead represent the ball as a circle using a single radius parameter, which is more suitable for spherical objects and reduces the parameter search space while providing a tighter geometric fit.
Crucially, unlike conventional object detectors that estimate only static object properties, our model jointly predicts both spatial and kinematic quantities by directly inferring 2D velocity vectors from the event representation. 
This leverages the temporal structure inherently encoded in the event stream, enabling accurate tracking of fast-moving objects without requiring multiple detection steps over time.

\textbf{Architecture}~The base architecture consists of a CSPDarknet53-based backbone for feature extraction~\cite{bochkovskiy2020yolov4}, followed by two detection heads operating at different scales with strides of 16 and 8. It uses anchor-based detection with two anchors per scale, corresponding to radii of 5, 10, and 20 pixels.
Each detection head outputs 6 channels per anchor box; 5 channels for regression targets $(x, y, r, v_x, v_y)$ and 1 channel for the objectness score.

\subsection{Trajectory Estimation with a Physics Model}
We use an \gls{ekf} to estimate the 3D ball state $\mathbf{x}(t) = [\mathbf{p}(t)\;\mathbf{v}(t)\;\boldsymbol{\omega}(t)]$, where $\mathbf{p}(t)$ denotes position, $\mathbf{v}(t)$ the velocity, and $\boldsymbol{\omega}(t)$ the spin.
The nonlinear continuous-time dynamics are given by $\dot{\mathbf{x}}(t) = f(\mathbf{x}(t))$, where $f(\cdot)$ encodes the ball motion model for ball aerodynamics and ball contact (table and racket contacts) as described in \cite{nakashima2011robotic}. 
Sensor measurements fed into the EKF are the triangulated position and velocity predictions, as well as spin estimates obtained from the \gls{aps} cameras (described in Section~\ref{sec:spin_estimation}).
To accurately predict the course of the ball trajectory it is critical to compute an optimal initial state.
We estimate the initial state using a batch nonlinear least squares fit over the first $N$ measurements.
For each measurement time $t_i$, the predicted state is obtained by forward propagation from the initial state $\mathbf{x}_0$ at time $t_0$ as $\hat{\mathbf{x}}_i(\mathbf{x}_0) = \Phi(\mathbf{x}_0, t_i - t_0)$, with corresponding predicted measurement $\hat{\mathbf{z}}_i(\mathbf{x}_0) = \mathbf{H} \hat{\mathbf{x}}_i(\mathbf{x}_0)$.
We define the weighted nonlinear least squares objective:
\begin{equation}
J(\mathbf{x}_0) =
\sum_{i=0}^{N-1}
\left(
\mathbf{z}_i - \hat{\mathbf{z}}_i(\mathbf{x}_0)
\right)^\top
\mathbf{R}_i^{-1}
\left(
\mathbf{z}_i - \hat{\mathbf{z}}_i(\mathbf{x}_0)
\right).
\end{equation}

where the optimal initial state is obtained from $\mathbf{x}_0^\star = \arg\min_{\mathbf{x}_0} J(\mathbf{x}_0)$.
This optimization problem is solved using the nonlinear least squares solver Levenberg–Marquardt.
After computing $\mathbf{x}_0^\star$, the EKF is not started directly at the last measurement. Instead, the filter is initialized at time $t_0$ and the first $N$ measurements are replayed using alternating prediction and update steps. This propagates the uncertainty through the nonlinear dynamics and leads to a consistent covariance matrix.
    
\section{Results}

We evaluate our pipeline along three complementary axes that mirror the requirements of human-robot rallies: detection accuracy, trajectory prediction quality, and end-to-end behavior on the robot.
We first quantify the 2D pixel accuracy of the \gls{cnn} detector, benchmarking it against prior event- and frame-based methods on both our dataset and existing datasets to isolate the gains from learning over hand-crafted detection.
We then turn to 3D trajectory prediction through the \gls{ekf}, where the key finding is that supplying the CNN's predicted image-plane velocity as an EKF measurement substantially reduces bounce-prediction error and timing variance compared to a position-only baseline; we further analyze how the detector update rate affects this accuracy.
Finally, we close the loop by reporting return rates from robot experiments.

\subsection{Experimental Setup}
\subsubsection{Neural Network Training}

As input to the network, event representations are constructed by accumulating \gls{sae}~\cite{lagorce2016hots} over a 5~ms window at the full sensor resolution of 1280 $\times$ 720. The resulting input consists of two channels, corresponding to the positive and negative event polarities.

\textbf{Data Augmentation}~To improve model robustness and generalization, we employ a set of geometric augmentations that preserve the physical consistency of ball position, radius, and velocity under spatial transformations.
Augmentations of translation, zoom, and flipping are applied independently and stochastically during training with probabilities of $p = 0.1$.
The augmentation strategy has been empirically validated to improve localization accuracy while reducing overfitting to specific camera viewpoints.

\textbf{Loss Scheduling}~The training loss is a weighted linear combination of a binary cross-entropy term for objectness and MSE terms for position, radius, and velocity.
We implement dynamic loss weight scheduling during training, allowing the network to initially focus on objectness and localization before emphasizing velocity prediction in later stages. This curriculum learning approach improves convergence and final performance, with weights interpolated linearly as $w_i(t) = w_i^{\text{start}} + \frac{t - t_{\text{start}}}{t_{\text{end}} - t_{\text{start}}}(w_i^{\text{end}} - w_i^{\text{start}})$.

\textbf{Optimization}~Training was conducted using the AdamW optimizer with a learning rate of $1\times10^{-4}$, momentum parameters $(0.9, 0.999)$, and $\epsilon = 10^{-8}$. 
We trained for up to 130k steps amounting to 30 epochs (approximately 12 hours) on a single NVIDIA GeForce RTX 4070 Ti GPU, with a batch size of 12 and two gradient accumulation steps.

\textbf{Training Data}~We subsample a part of the recorded dataset for training, as seen in Table~\ref{tab:dataset_train}.
We split the dataset so that all recordings from a completely separate laboratory, featuring a different camera setup and lighting configuration, are reserved exclusively for the test set, enabling evaluation of generalization.
In total, we use 13.5 minutes of data for training and validating our proposed model.

\begin{table}[tbp]
\centering
\caption{Statistics of the training data. An event frame corresponds to 5~ms of accumulated event data.}
\label{tab:dataset_train}
\setlength{\tabcolsep}{4pt}
\begin{tabular}{@{}lcccc@{}}
\toprule
\textbf{Split} & \textbf{Elite} & \textbf{Amateur} & \textbf{Lab(s)} & \textbf{Total} \\ \midrule
\textbf{Train} & 64.7k & 35.1k & A    & 99.7k  \\
\textbf{Test}  & 12.7k & 48.8k & A, B & 61.5k  \\ \midrule
\textbf{All}   & 77.4k & 83.9k &      & 161.2k \\ \bottomrule
\end{tabular}
\end{table}

\subsubsection{Real-Time Inference Pipeline}

To achieve high-frequency updates in our multi-camera system, we implement a C++ pipeline for parallel inference using TensorRT. 
On an NVIDIA GeForce RTX 4090, the model achieves an inference rate of 550 Hz using four parallel CUDA streams (corresponding to four cameras) at FP32 precision.
Switching to FP16 precision increases throughput to 1147~Hz without significant loss in accuracy.
The input tensors are generated in parallel across separate threads for each camera. However, these threads contend for DRAM bandwidth when copying \gls{sae}s to CUDA host memory, leading to bandwidth saturation.
By representing the input tensors directly in FP16 precision, we reduce memory bandwidth usage by 50\% and data transfer time by approximately 1 ms. 

\begin{table}[tbp]
\centering
\caption{End-to-end latency breakdown of the perception pipeline (ms).}
\label{tab:latency}
\setlength{\tabcolsep}{3pt}
\begin{tabular}{@{}ccccc|c@{}}
\toprule
\textbf{USB} & \textbf{\begin{tabular}[c]{@{}c@{}}Event frame\\ generation\end{tabular}} & \textbf{\begin{tabular}[c]{@{}c@{}}GPU\\ upload\end{tabular}} & \textbf{Inference} & \textbf{Triang.} & \textbf{Total} \\ \midrule
up to 6.0 & 1.9 & 2.0 & 1.8 & 0.8 & \textbf{up to 12.5} \\ \bottomrule
\end{tabular}
\end{table}

Overall, parallelizing the pipeline, distributing operations across multiple threads, and reducing precision enable the full camera-to-3D state estimation pipeline to run at 300 Hz with four cameras or 400 Hz with three cameras. 
Empirically, we observe that comparable spatial coverage can be achieved with three cameras when the third camera is placed at the center-left or center-right of the table, rather than in a corner configuration.
Based on this observation, we adopt a three-camera setup operating at 400 Hz, using a sliding window over 5 ms accumulated events to avoid restricting runtime performance.
Table~\ref{tab:latency} provides a breakdown of the system latency from initial event generation at the sensor to successful triangulation.

\subsubsection{Robot Platform}
\label{sec:robot_platform}
We integrate our event-based perception system into a robotic platform consisting of a St\"aubli TX2-60L robotic arm~\cite{staubli_tx2_60l}. 
A motion planner based on Ruckig~\cite{berscheid2021jerk} then generates a trajectory to the predicted interception point. Ruckig computes time-optimal trajectories from arbitrary initial states under velocity, acceleration, and jerk constraints.
The full pipeline from input processing to robot execution is illustrated in Figure~\ref{figures:robot_events}.

\subsection{2D Ball Detection Accuracy}


\begin{table*}[t]
\centering
\caption{Comparative evaluation of existing methods on the dataset provided by Ziegler et al.~\cite{ziegler2025event} and on our newly collected dataset. Values are reported as mean $\pm$ std, and those highlighted in bold denote the best performance.}
\label{tab:accuracy_comparison}
\begin{threeparttable}
\begin{tabular}{@{}lccccccc@{}}
\toprule
                                                                             & \multicolumn{1}{l|}{}              & \multicolumn{2}{c|}{\textbf{Ziegler et al. \cite{ziegler2025event} Data}}       & \multicolumn{3}{c|}{\textbf{Our Data}}                                                                                                     &                                                  \\ \midrule
\multicolumn{1}{l|}{\textbf{Method}}                                         & \multicolumn{1}{l|}{\textbf{Input}} & \textbf{Error (px) $\downarrow$} & \multicolumn{1}{c|}{\textbf{IoU $\uparrow$}} & \textbf{Error (px) $\downarrow$} & \textbf{IoU $\uparrow$} & \multicolumn{1}{c|}{\textbf{\begin{tabular}[c]{@{}c@{}}Detection \\ Rate $\uparrow$\end{tabular}}} & \textbf{Update Rate (Hz)}                        \\ \midrule
\multicolumn{1}{l|}{\textbf{Tebbe et al.~\cite{tebbe2019table}}}             & \multicolumn{1}{c|}{Frame}          & 1.43                             & \multicolumn{1}{c|}{**}                      & **                               & **                      & \multicolumn{1}{c|}{**}                                                                            & 149                                              \\
\multicolumn{1}{l|}{\textbf{Median filter~\cite{ziegler2025event}}}          & \multicolumn{1}{c|}{Event}          & 2.90 ± 1.60                      & \multicolumn{1}{c|}{*}                       & 3.13 ± 2.84                      & *                       & \multicolumn{1}{c|}{36.0~\%$^{\dagger}$}                                                          & \textbf{13750 ± 1196}                    \\
\multicolumn{1}{l|}{\textbf{Particle filter~\cite{gloverrobust}}}            & \multicolumn{1}{c|}{Event}          & 2.93 ± 1.13                      & \multicolumn{1}{c|}{0.55 ± 0.08}             & 3.10 ± 2.78                      & 0.63 ± 0.09             & \multicolumn{1}{c|}{36.0~\%$^{\dagger}$}                                                          & 3493 ± 403                               \\
\multicolumn{1}{l|}{\textbf{Ziegler et al.~\cite{ziegler2025event}}}         & \multicolumn{1}{c|}{Event}          & 1.34 ± 0.79                      & \multicolumn{1}{c|}{0.78 ± 0.11}             & 3.03 ± 2.54                      & 0.64 ± 0.09             & \multicolumn{1}{c|}{36.0~\%$^{\dagger}$}                                                                       & 4136 ± 101                               \\
\multicolumn{1}{l|}{\textbf{Alberico et al.~\cite{alberico2025egocentric}}} & \multicolumn{1}{c|}{Event}          & 2.54 ± 2.27                      & \multicolumn{1}{c|}{0.74 ± 0.26}             & 4.31 ± 4.22                      & 0.75 ± 0.19             & \multicolumn{1}{c|}{\textbf{98.2~\%}}                                                              & 200                                       \\
\multicolumn{1}{l|}{\textbf{Ours}}                                           & \multicolumn{1}{c|}{Event}          & \textbf{0.91± 0.54}              & \multicolumn{1}{c|}{\textbf{0.78 ± 0.09}}    & \textbf{0.75 ± 2.95}             & \textbf{0.80 ± 0.09}    & \multicolumn{1}{c|}{96.8~\%}                                                                       & 400                                              \\ \bottomrule
\end{tabular}
\begin{tablenotes}[flushleft]\footnotesize
\item[] \textsuperscript{*}Computation is not applicable. \quad \textsuperscript{**}Code is not available.
\item[$\dagger$] The median filter and particle filter baselines share the same \gls{roi} initialization, which causes them to fail on the same trajectories. Ziegler et al.~\cite{ziegler2025event} instead bootstrap with their method at full sensor resolution; the matching 36.0\% detection rate is therefore coincidental.
\end{tablenotes}
\end{threeparttable}
\end{table*}

\begin{table}[tbp]
\centering
\caption{Generalization of the detector to an unseen environment (mean $\pm$ std). Detection and image-plane velocity accuracy are compared on the in-distribution lab (A) and a held-out lab (B).}
\label{tab:generalization_comparison}
\begin{tabular}{@{}l|cccc@{}}
\toprule
\textbf{Lab} & \textbf{\begin{tabular}[c]{@{}c@{}}Localization \\ MAE (px) $\downarrow$ \end{tabular}} & \textbf{IoU} $\uparrow$ & \textbf{\begin{tabular}[c]{@{}c@{}}Velocity Mag. \\ MAE (px) $\downarrow$\end{tabular}} & {\color[HTML]{333333} \textbf{\begin{tabular}[c]{@{}c@{}}Velocity Ang. \\ (°) $\downarrow$ \end{tabular}}} \\ \midrule
\textbf{A} & 0.75 ± 1.21 & 0.79 ± 0.1 & 0.93 ± 0.81 & 12.37 ± 20.2 \\
\rowcolor[HTML]{EFEFEF} 
\textbf{B} & 0.76 ± 4.98 & 0.81 ± 0.09 & 0.95 ± 0.82 & 12.18 ±15.7 \\ \bottomrule
\end{tabular}
\end{table}

We benchmark our neural network against existing event-based methods for table tennis ball detection~\cite{ziegler2025event, gloverrobust, alberico2025egocentric}. 
For methods with update rates above 1~kHz, errors are computed only at the maximum available label rate.
Our method outperforms all prior approaches on the dataset introduced by Ziegler et al.~\cite{ziegler2025event}, 
achieving a \gls{mae} of 0.91~pixels and \gls{iou} of 0.78 (see Tab.~\ref{tab:accuracy_comparison}), 
without any fine-tuning of the network or alterations to the input events.

Our dataset proves considerably more challenging: existing model-based methods show a significant drop in accuracy relative to the results reported on the dataset of Ziegler et al., which contains just four trajectories and no scene activity beyond the ball.
These results highlight the increased diversity of our dataset, which includes background activity from players and a wide range of ball trajectories with varying speed.

We define the detection rate as the percentage of trajectories for which \gls{roi} initialization is successful for prior methods and report results only on successful \gls{roi} initializations to ensure a fair evaluation. 
For our method, the detection rate is defined as the percentage of detections whose confidence score exceeds an empirical threshold of 0.1.
For the comparison with Alberico et al.~\cite{alberico2025egocentric}, we set the \gls{roi} to the ground truth position instead of the gaze re-projection since gaze information is not applicable for our method.
This gives their detection rate in \ref{tab:accuracy_comparison} a slight advantage.
Initially, most of the clusters were rejected and we only achieved a detection rate of 8.2\%; however, after tuning the DBSCAN parameters, the detection rate increased to 96.8\%.
When evaluated on the dataset of Ziegler et al., we could not successfully improve the detection rate of 9.8\% by tuning the parameters and therefore left the parameters as they were set originally.
The low detection rates of the median filter, particle filter, and Ziegler et al. (failing on 64\% of trajectories) stem from the brittleness of their classical, model-based \gls{roi} initialization, which is sensitive to background motion from players and other scene clutter.

We further note that the update rates of all prior event-based methods are reported per single camera on CPU, whereas our 400~Hz corresponds to running three cameras on GPU. Despite this, our update rate still exceeds the single-camera CPU rates of Alberico et al. (200~Hz) and the frame-based pipeline of Tebbe et al.~\cite{tebbe2019table} (149~Hz). The remaining event-based baselines reach higher per-camera rates only by trading off accuracy or robustness.

In comparison to all previous approaches, including Tebbe et al. which uses a frame-based approach, our method reports the best \gls{mae} error of 0.75~px and 0.91~px reported on our dataset and the dataset of Ziegler et al., respectively.
For reference, a 2D error of 0.75~px translates to an \gls{mae} of approximately 8~mm in 3D space. 
This is computed using the pinhole camera model $e_{3D} = \frac{e_\text{pix}}{f} \, d^2$ with a focal length of 850~px and a 3~m distance to the ball.

The model maintains comparable performance across both environments, with only negligible differences in mean localization and velocity estimation metrics (see Tab.~\ref{tab:generalization_comparison}). 
These results indicate that the model generalizes well to unseen environments, although the higher variance suggests slight sensitivity to lab-specific factors.

\subsection{3D Trajectory Prediction Accuracy}

We evaluate our pipeline in 3D through \gls{ekf}-based trajectory prediction.
Our elite-level test set contains roughly 100 segments with a ball bounce on the table.
In all \gls{ekf} experiments, spin estimates obtained from the \gls{aps} cameras using the method described in Section~\ref{sec:spin_estimation} are used as additional measurements to initialize and update the spin component of the state vector.

\subsubsection{Velocity as an EKF Measurement}

\begin{table}[tbp]
\centering
\caption{Impact of velocity initialization on EVS-based bounce prediction accuracy (mean $\pm$ std).}
\label{tab:ekf_vel}
\begin{tabular}{@{}ccc@{}}
\toprule
\textbf{Method} & \textbf{Bounce RMSE (cm) $\downarrow$} & \textbf{Delta T (ms) $\downarrow$}     \\ \midrule
EVS pos + vel   & $7.7 \pm 16.5$  & $-3.3 \pm 23.2$ \\
EVS pos only    & $12.1 \pm 18.7$  & $-2.4 \pm 36$  \\ \midrule
\rowcolor[HTML]{EFEFEF} 
Pseudo GT (lower bound)  & $1.8 \pm 4.3$  & $2.6 \pm 9.4$  \\ \bottomrule
\end{tabular}
\end{table}

Our central trajectory-prediction result is that treating the CNN's predicted image-plane velocity as an EKF measurement substantially improves bounce prediction. Incorporating velocity reduces the mean bounce RMSE from 12.1~cm to 7.7~cm, a 4.4~cm (36\%) relative improvement, and it also tightens the spread of the bounce-timing error from $\pm 36$~ms to $\pm 23.2$~ms (see Tab.~\ref{tab:ekf_vel}).
For this ablation, we vary the initialization intervals from 6~ms to 20~ms and the time to bounce from 50~ms to 200~ms to enable a more comprehensive evaluation. 
The minimum initialization time of 6~ms ensures that at least two APS-based position estimates are available, enabling a non-zero velocity estimate during initialization. 
We compare our method against a baseline \gls{ekf} that uses pseudo ground-truth states obtained from the APS pipeline; higher EKF prediction accuracy with APS measurements is expected, as they are used as pseudo ground truth and define an empirical lower bound.

\subsubsection{Effect of Detector Update Rate}

\begin{figure}[tbp]
    \centering
    \includegraphics[width=\linewidth]{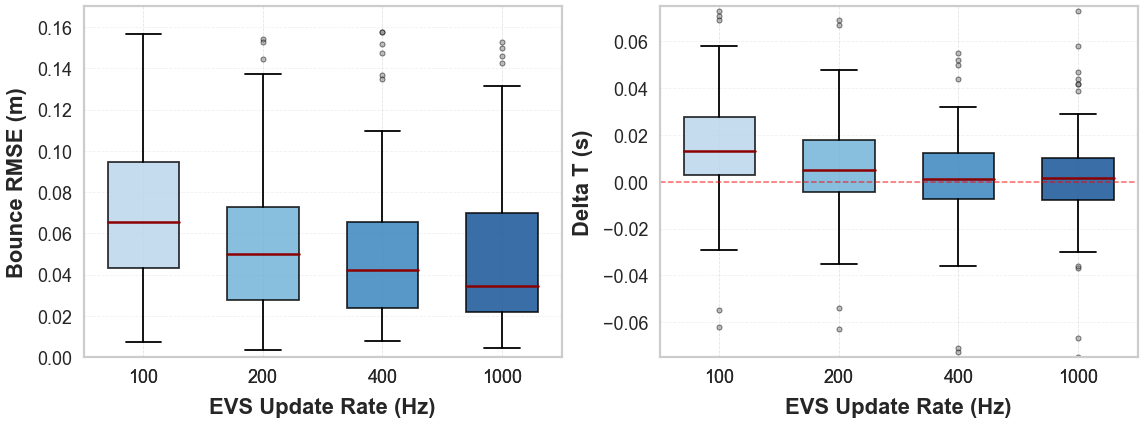}
    \caption{Effect of EVS update rate on EKF-based trajectory prediction, measured by bounce localization error (RMSE) and timing offset ($\Delta T$).}
    \label{figures:ekf_evs_rate_comp}
\end{figure}

We complement the velocity ablation by analyzing the effect of the \gls{evs} measurement update rate on \gls{ekf}-based ball bounce prediction, simulating different online update frequencies of the detection pipeline, ranging from 100 Hz to 1000 Hz. 
The prediction horizon (i.e., time to bounce) is fixed at 100~ms, with an initialization period of 10~ms to ensure a fair comparison, particularly for the 100~Hz case where at least two measurements are required.
Prediction accuracy improves with increasing observation update rate; however, the gains in estimating both the bounce location and timing diminish beyond 400 Hz (see Fig.~\ref{figures:ekf_evs_rate_comp}).
This suggests that our achieved online update rate of 400 Hz is near-optimal for the \gls{ekf}.

\subsection{Robot Experiments}
To validate the system described in Sec.~\ref{sec:robot_platform}, we first use a ball cannon with fixed velocity and direction. In this setting, the robot achieves a return rate of 100\% (35/35 successful returns). We further evaluate the system in human–robot rallies, achieving a return rate of 75\%. Supporting videos for both experiments are provided in the supplementary video.

\section{Conclusions}
In this work, we present the first fully integrated EVS-based perception system enabling human–robot table tennis rallies. 
For robust online state estimation at high frequency and accuracy, we train a \gls{cnn} on our dataset and demonstrate that it outperforms previous methods in ball detection accuracy, thereby introducing the first learning-based approach for event-based ball position estimation in table tennis robotics. 
Training such a model is enabled by the first large-scale event-camera dataset for table tennis, which we release alongside this work: over 1000 rallies and 5~h~10~min of play, recorded across players ranging from amateurs to elite athletes. 
In addition, we exploit the high temporal resolution of event cameras to predict the ball's velocity directly in the image plane using our \gls{cnn}-based approach.
We show that incorporating velocity as a sensor measurement enables an \gls{ekf} to achieve significantly improved trajectory prediction compared to using position-only measurements, reducing 100~ms bounce-prediction error by 36\% (12.1~cm $\rightarrow$ 7.7~cm) and tightening bounce-timing variance from $\pm 36$~ms to $\pm 23.2$~ms.
Overall, this work demonstrates the potential of event-based vision for high-speed sports scenarios and marks an important step toward event-based sensing in robotic sports applications.

Looking ahead, an opportunity lies in fully exploiting the microsecond temporal resolution of event cameras. Our current implementation already delivers real-time performance on commodity GPUs, providing a strong reference point for the accuracy achievable with event-frame representations. 
Asynchronous, event-driven processing on neuromorphic hardware is a natural next step: as such platforms mature, they offer a promising path toward closing the gap between sensor-level and system-level latency.

\section{Acknowledgments}
We thank Andreas Ziegler for sharing their dataset and code, and Valentin Monferrato and Adriano Patan\`e for their support in dataset labeling.

\bibliographystyle{IEEEtran}
\bibliography{bib}

\end{document}